\documentclass[10pt,twocolumn,letterpaper]{article}	

\usepackage{cvpr}
\usepackage{times}
\usepackage{epsfig}
\usepackage{graphicx}
\usepackage{amsmath}
\usepackage{amssymb}
\usepackage{graphicx}
\usepackage{subcaption}
\usepackage{setspace}

\usepackage{calrsfs}
\DeclareMathAlphabet{\pazocal}{OMS}{zplm}{m}{n}

\newcommand{\Ib}{\pazocal{I}}	
\newcommand{\Vb}{\pazocal{V}}

\usepackage[margin=2cm]{geometry}
\usepackage{adjustbox}
\usepackage{multirow}
\usepackage{booktabs}

\let\svthefootnote\thefootnote

\newcommand\blankfootnote[1]{%
  \let\thefootnote\relax\footnotetext{#1}%
  \let\thefootnote\svthefootnote%
}
\let\svfootnote\footnote
\renewcommand\footnote[2][?]{%
  \if\relax#1\relax%
    \blankfootnote{#2}%
  \else%
    \if?#1\svfootnote{#2}\else\svfootnote[#1]{#2}\fi%
  \fi
}

\usepackage[pagebackref=true,breaklinks=true,letterpaper=true,colorlinks,bookmarks=false]{hyperref}

\cvprfinalcopy 

\def\cvprPaperID{0451} 

\begin{document}
\title{Temporal 3D ConvNets:\\
New Architecture and Transfer Learning for Video  Classification}
\author{
    Ali Diba$^{1,4,\star}$, Mohsen Fayyaz$^{2,\star}$, Vivek Sharma$^{3}$, Amir Hossein Karami$^{4}$, Mohammad Mahdi Arzani$^{4}$, \\ Rahman Yousefzadeh$^{4}$, Luc Van Gool$^{1,4}$\\
    {\normalsize $^{1}$ESAT-PSI, KU Leuven, $^{2}$University of Bonn, $^{3}$CV:HCI, KIT, Karlsruhe, $^{4}$Sensifai} \\ 
     \tt\small \{firstname.lastname\}@esat.kuleuven.be, \{lastname\}@sensifai.com,  \\ \tt\small fayyaz@iai.uni-bonn.de,  vivek.sharma@kit.edu 
 }


\maketitle

\footnote[]{$^{\star}$Ali Diba and Mohsen Fayyaz contributed equally to this work. Mohsen Fayyaz contributed to this work while he was at Sensifai.}

\begin{abstract}

The work in this paper is driven by the question how to exploit the temporal cues available in videos for their accurate classification, and for human action recognition in particular? Thus far, the vision community has focused on spatio-temporal approaches with fixed temporal convolution kernel depths. We introduce a new temporal layer that models variable temporal convolution kernel depths. We embed this new temporal layer in our proposed 3D CNN. We extend the DenseNet architecture - which normally is 2D - with 3D filters and pooling kernels. We name our proposed video convolutional network `Temporal 3D ConvNet'~(T3D) and its new temporal layer `Temporal Transition Layer'~(TTL). Our experiments show that T3D outperforms the current state-of-the-art methods on the HMDB51, UCF101 and Kinetics datasets.

The other issue in training 3D ConvNets is about training them from scratch with a huge labeled dataset to get a reasonable performance. So the knowledge learned in 2D ConvNets is completely ignored. Another contribution in this work is a simple and effective technique to transfer knowledge from a pre-trained 2D CNN to a randomly initialized 3D CNN for a stable weight initialization. This allows us to significantly reduce the number of training samples for 3D CNNs. Thus, by finetuning this network, we beat the performance of generic and recent methods in 3D CNNs, which were trained on large video datasets, e.g. Sports-1M, and finetuned on the target datasets, e.g. HMDB51/UCF101. The T3D codes will be released soon\footnote{\url{https://github.com/MohsenFayyaz89/T3D}}. 

\end{abstract}



\section{Introduction} \label{sec:intro}

Compelling advantages of exploiting temporal rather than merely spatial cues for video classification have been shown lately~\cite{tle,c3d,n3d}. Such insights are all the more important given the surge in multimedia videos on the Internet. Even if considerable progress in exploiting temporal cues was made~\cite{i3d,pooling,c3d,res3d}, the corresponding systems are still wanting. Recently, several variants of Convolutional Neural Networks (ConvNets) have been proposed that use 3D convolutions, but they fail to exploit long-range temporal information, thus limiting the performance of these architectures. Complicating aspects include: (i) these video architectures have many more parameters than 2D ConvNets; (ii) training the video architectures calls for extra large labeled datasets; and (iii) extraction and usage of optical-flow maps which is very demanding, and also difficult to obtain for large scale dataset, e.g. Sports-1M. All of these issues negatively influence their computational cost and performance. Two ways to avoid these limitations are (i) an architecture that efficiently captures both appearances and temporal information throughout videos, thus avoiding the need of optical-flow maps; and (ii) an effective supervision transfer that bridges the knowledge transfer between different architectures, such that training the networks from scratch is no longer needed.

\begin{figure*}[ht]
 \centering
 \includegraphics[width=2\columnwidth]{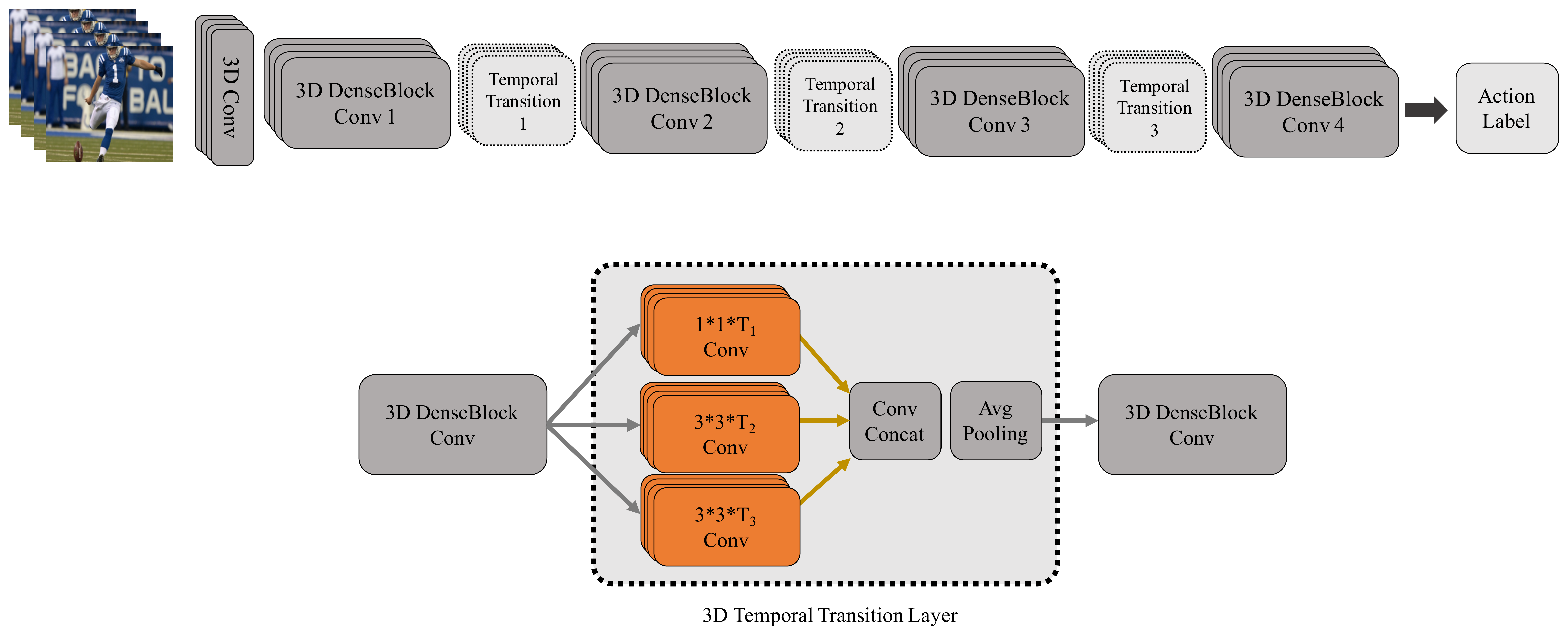}
 \caption{\textbf{Temporal 3D ConvNet~(T3D).} Our Temporal Transition Layer~(TTL) is applied to our DenseNet3D. T3D uses video clips as input. The 3D feature-maps from the clips are densely propagated throughout the network. The TTL operates on the different temporal depths, thus allowing the model to capture the appearance and temporal information from the short, mid, and long-range terms.  The output of the network is a video-level prediction.} 
  \label{fig:3dnet}
 \end{figure*}

Motivated by the above observations, we introduce a novel deep spatio-temporal feature extractor network illustrated in Figure~\ref{fig:3dnet}. The aim of this extractor is to model variable temporal 3D convolution kernel depths over shorter and longer time ranges. We name this new layer in 3D ConNets configuration `Temporal Transition Layer'~(TTL). TTL is designed to concatenate temporal feature-maps extracted at different temporal depth ranges, rather than only considering fixed 3D homogeneous kernel depths~\cite{i3d,c3d,res3d}. We embed this new temporal layer into the 3D CNNs. In this work, we extend the DenseNet architecture - which by default has 2D filters and pooling kernels - to incorporate 3D filters and pooling kernels, namely DenseNet3D. We used DenseNet because it is highly parameter efficient.  Our TTL replaces the standard transition layer in the DenseNet architecture. We refer to our modified DenseNet architecture as `Temporal 3D ConvNets'~(T3D), inspired by C3D~\cite{c3d}, Network in Network~\cite{nin}, and DenseNet~\cite{densenet} architectures. T3D densely and efficiently captures the appearance and temporal information from the short, mid, and long-range terms. We show that the TTL feature representation fits action recognition well, and that it is a much simpler and more efficient representation of the temporal video structure. The TTL features are densely propagated throughout the T3D architecture and are trained end-to-end. In addition to achieving high performance, we show that the T3D architecture is computationally efficient and robust in both the training and inference phase. T3D is evaluated on three challenging action recognition datasets, namely HMDB51, UCF101, and Kinetics. We experimentally show that T3D achieves the state-of-the-art performance on HMDB51 and UCF101 among the other 3D ConvNets and competitive results on Kinetics. 
 
It has been shown that training 3D ConvNets~\cite{c3d} from scratch takes two months~\cite{res3d} for them to learn a good feature representation from a large scale dataset like Sports-1M, which is then finetuned on target datasets to improve performance. Another major contribution of our work therefore is to achieve supervision transfer across architectures, thus avoiding the need to train 3D CNNs from scratch. Specifically, we show that a 2D CNN pre-trained on ImageNet can act as `\textit{a teacher}' for supervision transfer to a randomly initialized 3D CNN for a stable weight initialization. In this way we avoid the excessive computational workload and training time. Through this transfer learning, we outperform the performance of generic 3D CNNs (C3D~\cite{c3d}) which was trained on Sports-1M and finetuned on the target datasets, HMDB51/UCF101.

The rest of the paper is organized as follows. In Section~\ref{sec:related}, we discuss related work. Section~\ref{sec:method} describes our proposed approaches. The implementation details,  experimental results and their analysis are presented in Section~\ref{sec:experiments}. Finally, conclusions are drawn in Section~\ref{sec:conclusion}.

\section{Related Work} \label{sec:related}

\noindent
\textbf{Video Classification with and without ConvNets:}
Video classification and understanding has always been a hot topic. Several techniques have been proposed to come up with efficient spatio-temporal feature representations that capture the appearance and motion propagation across frames in videos, such as HOG3D~\cite{ref15}, SIFT3D~\cite{ref23}, HOF~\cite{ref18}, ESURF~\cite{ref38}, MBH~\cite{ref4}, iDTs~\cite{ref34}, and more. These were all hand-engineered. Among these, iDTs yielded the best performance, at the expense of being computationally expensive and lacking scalability to capture semantic concepts. It is noteworthy that recently several other techniques~\cite{rankbasura} have been proposed that also try to model the temporal structure in an efficient way.


\textbf{Temporal ConvNets:} Recently, several temporal architectures have been proposed for video classification, where the input to the network consists of either RGB video clips or stacked optical-flow frames. The filters and pooling kernels for these architectures are 3D (x, y, time) with fixed temporal kernel depths throughout the architecture. The most intuitive are 3D convolutions~($s \times s\times d$)~\cite{n3d} where the kernel temporal depth $d$ corresponds to the number of frames used as input, and $s$ is the kernel spatial size. Simonyan et al.~\cite{twostream} proposed a two-stream network, cohorts of RGB and flow ConvNets. In their flow stream ConvNets, the 3D convolution has $d$ set to 10. Tran et al.~\cite{c3d} explored 3D ConvNets with filter kernel of size $3\times 3 \times 3$. Tran et al. in~\cite{res3d} extended the ResNet architecture with 3D convolutions. Feichtenhofer et al.~\cite{pooling} propose 3D pooling. Sun et al.~\cite{sun3d} decomposed the 3D convolutions into 2D spatial and 1D temporal convolutions. Carreira et al.~\cite{i3d} proposed converting a pre-trained the 2D Inception-V1~\cite{googlenet} architecture to 3D by inflating all the filters and pooling kernels with an additional temporal dimension $d$. They achieve this by repeating the weights of 2D filters $d$ times for weight initialization of 3D filters. All these architectures have fixed temporal 3D convolution kernel depths throughout the whole architecture. To the best of our knowledge, our architecture is the first end-to-end deep network that integrates variable temporal depth information over shorter and longer temporal ranges.

\textbf{Transfer Learning:}  Finetuning or specializing the learned feature representations of a pre-trained network trained on another dataset to a target dataset is commonly referred to as transfer learning. Recently, several works have shown that transferring knowledge within or across modalities (e.g. RGB$\rightarrow$RGB~\cite{hintonrgb} vs. RGB$\rightarrow$Depth~\cite{judyrgbd}, RGB$\rightarrow$Optical-Flow~\cite{judyrgbd,alirgbo}, RGB$\rightarrow$Sound~\cite{l3}, Near-Infrared$\rightarrow$RGB~\cite{rgbir}) is effective, and leads to significant improvements in performance. They typically amount to jointly learning representations in a shared feature space. Our work differs substantially in scope and technical approach. Our goal is to transfer supervision across architectures (i.e. 2D$\rightarrow$3D ConvNets), not necessarily limited to transferring information between RGB models only, as our solution can be easily adopted across modalities too.

\section{Proposed Method} \label{sec:method}


Our goal is to capture short, mid, and long term dynamics for a video representation that embodies more semantic information. We propose a Temporal Transition Layer (TTL) inspired by GoogLeNet~\cite{googlenet}. It consists of several 3D Convolution kernels, with diverse temporal depths (see Fig.~\ref{fig:3dnet}). The TTL output feature maps are densely fed forward to all subsequent layers, and are learned end-to-end, as shown in Fig.~\ref{fig:3dnet}. We employ this TTL layer in a DenseNet3D architecture. We name the resulting networks as Temporal 3D ConvNets~(T3D). Fig.~\ref{fig:3dnet} sketches the steps of the proposed T3D. In addition, another major contribution of our work, we show supervision and knowledge transfer between cross architectures (i.e. 2D$\rightarrow$3D ConvNets), thus avoiding the need to train 3D ConvNets from scratch. More details about the transfer learning is given in Section~\ref{subsec:tranferlearning}.

\subsection{Temporal 3D ConvNets}

In this work, we use the DenseNet architecture which has 2D filters and pooling kernels, but extend it with 3D filters and pooling kernels. We used the DenseNet architecture for several reasons, such as simpler and highly parameter efficient deep architecture, its dense knowledge propagation, and state-of-the-art performance on image classification tasks. In specific, (i) we modify 2D DenseNet by replacing the 2D kernels by 3D kernels in the standard DenseNet architecture and we present it as DenseNet3D; and (ii) introducing our new Temporal 3D ConvNets~(T3D) by deploying 3D temporal transition layer~(TTL) instead of transition layer in DenseNet. In both setups, the building blocks of the network and the architecture choices proposed in~\cite{densenet} are kept same.

\textbf{Notation.} The output feature-maps of the 3D Convolutions and pooling kernels at the $l^{th}$ layer extracted for an input video, is a matrix $x \in \mathbb{R}^{h\times w \times c}$ where $h$, $w$ and $c$ are the height, width, and number of channels of the feature maps, resp. The 3D convolution and pooling kernels are of size (${s\times s \times d}$), where $d$ is the temporal depth and $s$ is the spatial size of the kernels. 

\textbf{3D Dense Connectivity.}
Similar to 2D dense connectivity, in our network it is 3D dense connectivity that directly connects the 3D output of any layer to all subsequent layers in the 3D Dense block. The composite function $H_{l}$ in the $l^{th}$ layer receives the $\{x_{i}\}_{i=0}^{l-1}$ 3D feature maps of all preceding  ($l-1$) layers as input. The output feature-map of $H_{l}$ in the $l^{th}$ layer is given by:  
\begin{equation}
x_{l}=H_{l}([x_{0},x_{1},\ldots,x_{l-1}])
\end{equation}
where [$x_{0},x_{1},\ldots,x_{l-1}$] denotes that the features maps are concatenated. The spatial sizes of the $x_{i}$ features maps are the same. The $H_{l}(\cdot)$ is a composite function of BN-ReLU-3DConv operations. 

\textbf{Temporal Transition Layer.}
Fig.~\ref{fig:3dnet} shows a sketch of Temporal Transition Layer~(TTL).
TTL is composed of several variable 3D Convolution temporal depth kernels and a 3D pooling layer, the depth of 3D Conv kernels ranges between $d, d \in \{T_{1},\ldots,T_{D}\}$, where $T_{d}$ have different temporal depths. The advantage of TTL is that it captures the short, mid, and long term dynamics, that embody important information not captured when working with some fixed temporal depth homogeneously throughout the network. 
The feature-map of $l^{th}$ layer is fed as input to the TTL layer, $TTL: x \rightarrow x^{'}$, resulting in a dense-aggregated feature representation $x^{'}$, where $x \in \mathbb{R}^{h \times w \times c}$ and $x^{'} \in \mathbb{R}^{h \times w \times c^{'}}$ . In specific, the feature-map from $l^{th}$, $x_{l}$ is convolved with $K$ variable 3D convolution kernel temporal depths, resulting to intermediate feature-maps $\{S_{1},S_{2},\ldots,S_{K}\}$, $S_{1}\in\mathbb{R}^{h \times w \times c_{1}}$,  $S_{2}\in\mathbb{R}^{h \times w \times c_{2}}$, $S_{K}\in\mathbb{R}^{h \times w \times c_{K}}$,  where $c_{1}$, $c_{2}$, and $c_{K}$ have different channel-depths as $x_{l}$ is convolved with different 3D convolution kernel temporal depths, while the spatial size ($h,w$) is same for all the $\{S_{k}\}_{k=1}^{K}$ feature-maps. These feature-maps $\{S_{k}\}_{k=1}^{K}$ are simply concatenated into a single tensor $[S_{1},S_{2},\ldots,S_{K}]$ and then fed into the 3D pooling layer, resulting to the output TTL feature-map $x^{'}$. The output of TTL, $x^{'}$ is fed as input to $(l+1)^{th}$ layer in the T3D architecture.  The TTL layer is learned in an end-to-end network learning, as shown in Fig.~\ref{fig:3dnet}.

\begin{figure*}[ht]
 \centering
 \includegraphics[width=2\columnwidth]{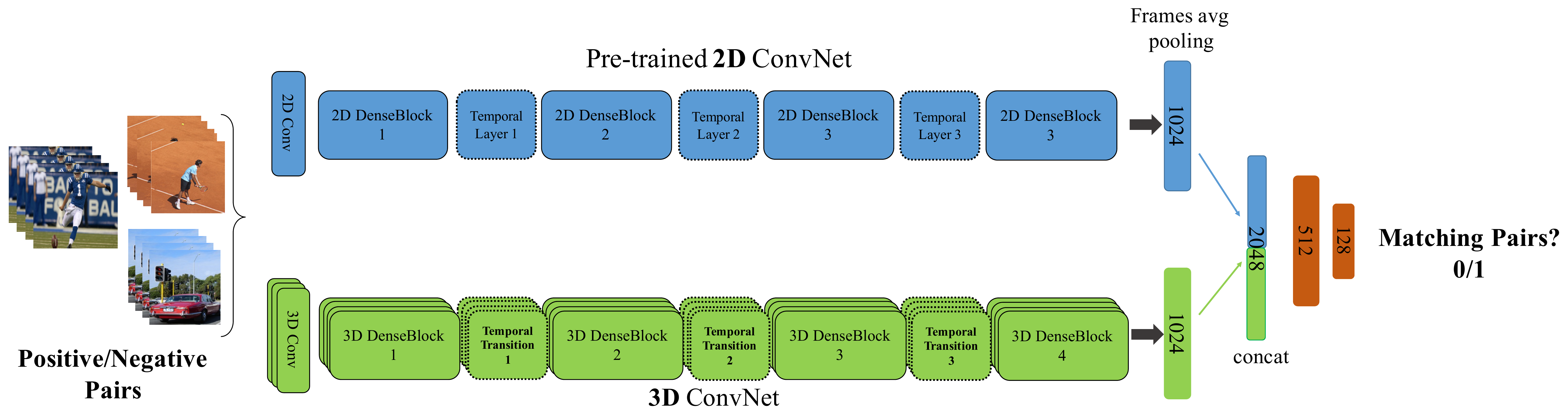}
 \caption{\textbf{Architecture for knowledge transfer from a pre-trained 2D ConvNet to 3D ConvNet.}  The 2D network operates on RGB frames, and the 3D network operates on video clips for the same time stamp. The 2D ConvNet acts as teacher for supervision transfer to 3D ConvNet, by teaching the 3D ConvNet to learn mid-level feature representation by image-video correspondence task. The model parameters of 2D ConvNet is frozen, while the task is to effectively learn the model parameters of 3D ConvNet only.} 
  \label{fig:transfer}
 \end{figure*}

In our work, we also compare T3D with DenseNet3D i.e with the standard transition layer but in 3D. Compared to the DenseNet3D, T3D performs significantly better in performance, shown in Experimental Section~\ref{sec:experiments}. Although we agree that T3D model has 1.3 times more model parameters than DenseNet3D, but it is worth to have it because of its outstanding performance. It is also worth saying that, one can readily employ our TTL in other architectures too such as in Res3D~\cite{res3d} or I3D~\cite{i3d}, instead of using fixed 3D Convolutions homogeneously through out the network.

\subsection{Supervision or Knowledge Transfer} \label{subsec:tranferlearning}

In this section, we describe our method for supervision transfer between cross architectures, i.e. pre-trained 2D ConvNets to 3D ConvNets, thus avoiding the need to train 3D ConvNets from scratch.

Lets assume we have a pre-trained 2D ConvNet $\Ib$ which has learned a rich representation from images, while $\Vb$ being a 3D ConvNet which is  randomly initialized using~\cite{densenetweighting} and  we want to transfer knowledge of rich representation from $\Ib$ to $\Vb$ for a stable weight initialization. This allows us to avoid training $\Vb$ from scratch which has million of more parameters, that requires heavy computational workload and training time of months~\cite{res3d}. In the current setup, $\Ib$ acts as a teacher for supervision transfer to $\Vb$ architecture.

Intuitively, our method uses correspondence between frames and video clips available by the virtue of them appearing together at the same time. Given a pair of $X$ frames and video clip for the same time stamp, the visual information in both frames and video are same. We leverage this for learning mid-level feature representation by image-video correspondence task between 2D and 3D ConvNet architectures, depicted in Figure~\ref{fig:transfer}. We use a pre-trained on ImageNet~\cite{imagenet}, 2D DenseNet CNN~\cite{densenet} as $\Ib$, and our T3D network as $\Vb$.  The 2D DenseNet CNN has 4 DenseBlock convolution layers and one fully connected layer at the end, while our 3D architecture as 4 3D-DenseBlocks and we add a fully-connected layer right after 3D Dense Block: Layer-4.  We simply concatenate the last $fc$ layers of both architectures, and connect them  with the \textbf{2048-dimensional fc} layer which is in turn connected to two fc layers with 512 and 128 sizes (fc1 , fc2) and to the final binary classifier layer.  We use a simple binary (0/1) matching classifier: given $X$ frames and a video clip - decide whether the pairs belong to each other or not.  For a given paired $X$-images and its corresponding video clip, the precise steps follows as,  $X$ frames are fed sequentially into the $\Ib$ and we average the $X$ last 2D fc features, resulting into \textbf{1024-D} feature representation, in parallel the video clip is fed to the $\Vb$, and we extract the 3D fc features (1024-D), and concatenate them, which is then passed to \textbf{fc1-fc2} for classification.

During the training, the model parameters of $\Ib$ is frozen, while the task is to effectively learn the model parameters of $\Vb$ without any additional supervision than correspondence between frames-video. The pairs belonging to same timestamp from the same video is a positive pairs, while the pairs coming from two different videos by random sampling of $X$ frames and video clip from two different videos is a negative pair. Note that, during back-propagation, only the model parameters for $\Vb$ is updated i.e. transferring the knowledge from $\Ib$ to $\Vb$. We have clearly shown in our experiments that a stable weight initialization of $\Vb$ is achieved, and when fine-tuned on the target dataset, allows the model to adapt quickly to the target dataset, thus avoiding training the model from scratch with improved performance. Also we proved that by using our proposed knowledge transfer method, 3D ConvNets can be trained directly on small dataset like UCF101 and achieve better performance than training from scratch.

To train the method, we used approx. 500K unlabeled video clips. Since our transfer learning is unsupervised and there is no need of video label. Further, our experiments in Section~\ref{sec:experiments} demonstrate that our proposed transfer learning of T3D outperforms the generic 3D ConvNets by a significant margin which was trained on large video dataset, Sports-1M~\cite{sport1m} and finetuned on the target datasets, HMDB51/UCF101.

\begin{table*}
\begin{center}
\resizebox{17.5cm}{!} {
\begin{tabular}{c|c|c|c|c}
\cline{1-5}
 {Layers}         &  {Output Size}      & DenseNet3D-121 & {T3D-121}      & {T3D-169}     \\                
\hline \hline
 3D Convolution    & $112\times112\times16$    & \multicolumn{3}{c}{$7\times7\times3$ conv, stride 2}     \\ \hline
 3D Pooling        & $56\times56\times16$      & \multicolumn{3}{c}{$3\times3\times3$ max pool, stride 1} \\ \hline
 
 3D Dense Block~(1)   & $56\times56\times16$	 & {$\begin{bmatrix} 1\times1\times1 \\ 3\times3\times3 \end{bmatrix}\times 6$ conv} & {$\begin{bmatrix} 1\times1\times1 \\ 3\times3\times3 \end{bmatrix}\times 6$ conv} & {$\begin{bmatrix} 1\times1\times1 \\ 3\times3\times3 \end{bmatrix}\times 6$ conv} \rule{0pt}{0.5ex} \\ \hline
 
  Transition/TTL			& $56\times56\times16$	 & $1\times 1\times1$ conv & $\begin{bmatrix} 1\times1\times\textbf{1} \\ 3\times3\times\textbf{3}  \\ 3\times3\times\textbf{6}  \end{bmatrix}$ conv $\rightarrow$ concat  & $\begin{bmatrix} 1\times1\times\textbf{1} \\ 3\times3\times\textbf{3}  \\ 3\times3\times\textbf{6}  \end{bmatrix}$ conv $\rightarrow$ concat \\
  \cline{2-5}
\rule{0pt}{1ex}  (1)			& $28\times28\times8$	 & \multicolumn{3}{c}{$2\times2\times2$ avg pool, stride 2}     \\ 
\hline

 3D Dense Block~(2)    & $28\times28\times8$	 & {$\begin{bmatrix} 1\times1\times1 \\ 3\times3\times3 \end{bmatrix}\times 12$ conv} & {$\begin{bmatrix} 1\times1\times1 \\ 3\times3\times3 \end{bmatrix}\times 12$ conv} & {$\begin{bmatrix} 1\times1\times1 \\ 3\times3\times3 \end{bmatrix}\times 12$ conv} \rule{0pt}{0.5ex} \\ \hline
  
  Transition/TTL			& $28\times28\times8$	 & $1\times 1\times1$ conv & $\begin{bmatrix} 1\times1\times\textbf{1} \\ 3\times3\times\textbf{3}  \\ 3\times3\times\textbf{4}  \end{bmatrix}$ conv   $\rightarrow$ concat & $\begin{bmatrix} 1\times1\times\textbf{1} \\ 3\times3\times\textbf{3}  \\ 3\times3\times\textbf{4}  \end{bmatrix}$ conv   $\rightarrow$ concat  \\
  \cline{2-5}
\rule{0pt}{1ex}  (2)			& $14\times14\times4$	 & \multicolumn{3}{c}{$2\times2\times2$ avg pool, stride 2}     \\ 
\hline

 3D Dense Block~(3)    & $14\times14\times4$	 & {$\begin{bmatrix} 1\times1\times1 \\ 3\times3\times3 \end{bmatrix}\times 24$ conv} & {$\begin{bmatrix} 1\times1\times1 \\ 3\times3\times3 \end{bmatrix}\times 24$ conv} & {$\begin{bmatrix} 1\times1\times1 \\ 3\times3\times3 \end{bmatrix}\times 32$ conv} \rule{0pt}{0.5ex} \\ \hline
  
 Transition/TTL			& $14\times14\times4$	 & $1\times 1\times1$ conv & $\begin{bmatrix} 1\times1\times\textbf{1} \\ 3\times3\times\textbf{3}  \\ 3\times3\times\textbf{4}  \end{bmatrix}$ conv   $\rightarrow$  concat & $\begin{bmatrix} 1\times1\times\textbf{1} \\ 3\times3\times\textbf{3}  \\ 3\times3\times\textbf{4}  \end{bmatrix}$ conv   $\rightarrow$  concat   \\
  \cline{2-5}
\rule{0pt}{1ex}  (3)			& $7\times7\times2$	 & \multicolumn{3}{c}{$2\times2\times2$ avg pool, stride 2}     \\ 
\hline

 3D Dense Block~(4)    & $7\times7\times2$	 & {$\begin{bmatrix} 1\times1\times1 \\ 3\times3\times3 \end{bmatrix}\times 16$ conv} & {$\begin{bmatrix} 1\times1\times1 \\ 3\times3\times3 \end{bmatrix}\times 16$ conv} & {$\begin{bmatrix} 1\times1\times1 \\ 3\times3\times3 \end{bmatrix}\times 32$ conv} \rule{0pt}{0.5ex} \\ \hline
  
Classification  & $1\times1\times1$			& \multicolumn{3}{c}{$7\times7\times2$ avg pool}  \\  
  \cline{2-5}
  Layer			&  & \multicolumn{3}{c}{400D softmax}     \\ 
\hline

\end{tabular}}
\end{center} \vspace{-0.3cm}
\caption{\textbf{Temporal 3D ConvNets~(T3D) Architectures.} All the proposed architectures incorporate 3D filters and pooling kernels. Each ``conv" layer shown in the table corresponds the composite sequence BN-ReLU-Conv operations.  The bold numbers shown in the TTL layer, denotes to the variable temporal convolution kernel depths applied to the 3D feature-maps.}
\label{table:arch}
\end{table*}

\section{Experiments} \label{sec:experiments}

In this section, we demonstrate a search for the architecture of the proposed T3D model, and then the configurations for input data. Afterwards, we first introduce the datasets and implementation details of our proposed approach. Following, we test and compare our proposed methods with baselines and other state-of-the-art methods. Finally, we compare our transfer learning: $2D\rightarrow3D$ ConvNet performance with generic state-of-the-art 3D CNN methods.  For the ablation study of architecture search and configurations of input data,  we  report  the  accuracy  of split 1 on UCF101.

\subsection{Architecture Search}
To find the best architecture for our T3D, we conduct a large scale architecture search. We start the search by designing a new DenseNet3D based on the 2D DenseNet architecture, then we explore T3D architecture based on our DenseNet3D. Due to the high computational time of 3D ConvNets we limit the exploring space by exploiting a lot of insights about good architectures~\cite{i3d,res3d}.

\paragraph{\textbf{DenseNet3D:}}
As mentioned before, we have designed a new DenseNet3D architecture. To achieve the best configuration for the new architecture we have done a series of tests on the network-size, and temporal-depth of input data to the network. For the architecture study, the model weights were initialized using~\cite{densenetweighting}.

We employ two versions of 2D-DenseNet with network sizes of 121 and 169 for designing the DenseNet3D, namely T3D-121 and T3D-169. Evaluations results of these two T3D models are reported in the Table~\ref{table:modelDepth}.

\begin{table}[hbt] 
\begin{center}
\resizebox{4cm}{!} {
\begin{tabular}{ c |   c}
\hline
Model Depth&   Accuracy~\%\\
\hline
\hline
121  		& 69.1\\
169  		& 71.3\\
\hline
\end{tabular}}
\end{center}\vspace{-0.3cm}
\caption{Evaluation results of DenseNet3D model with network sizes of 121 and 169 on UCF101 split 1. All models were trained from scratch.}
\label{table:modelDepth}
\end{table}

Temporal depth of series of input frames plays a key role in activity recognition tasks. Therefore, we have evaluated our T3D with configurations for different temporal depths. In the Table~\ref{table:temporalDepth} we have reported the performance of these tests.

\begin{table}[hbt] 
\begin{center}
\resizebox{4.5cm}{!} {
\begin{tabular}{ c |   c}
\hline
Temporal Depth&   Accuracy~\% \\
\hline
\hline
16  		& 66.8\\
32  		& 69.1\\
\hline
\end{tabular}}
\end{center}\vspace{-0.3cm}
\caption{Evaluation results of DenseNet3D model with temporal depths of 16 and 32 on UCF101 split 1. All models were trained from scratch.}
\label{table:temporalDepth}
\end{table}

\paragraph{\textbf{T3D:}}
Seeking for the best configurations of our T3D, we have done experiments based on the results of experimental study which are exploited from DenseNet3D. The TTL layer is added to make more efficient spatial-temporal connection between 3D DenseBlocks of convolutions, and improvement is observed in the performance. TTL models extract more informative spatial and temporal features because of the variable temporal convolution kernel sizes. The details of TTL layer are shown in Table~\ref{table:arch}. The growth rate of 32 is used for DenseNet3D and T3D, which indicates number of filters added to each layer of convolutions in DenseBlocks.

\paragraph{\textbf{T3D v.s Inception/ResNet 3D:}} For a fair comparison with other state-of-the-art architectures, we also implemented 3D ConvNets based on Inception~\cite{googlenet} and ResNet-34~\cite{resnet} architectures. This allows us to have deeper insights of other architectures performance. We believe all the models have almost same level of capacity to be generalized to 3D ConvNets, further a more smarter way to exploit the temporal information is an add-on, that is what our TTL provides, this can be clearly seen in Table~\ref{table:TTL}.

\begin{table}[hbt] 
\begin{center}
\resizebox{6cm}{!} {
\begin{tabular}{ c |   c}
\hline
3D ConvNet&   Accuracy~\%\\
\hline
\hline
ResNet3D-50  		& 59.2\\
Inception3D  		& 69.5\\
DenseNet3D-121 (ours)  		& 69.1\\
\textbf{T3D (ours)}     & \textbf{71.4}\\
\hline
\end{tabular}}
\end{center}\vspace{-0.3cm}
\caption{Evaluation results of T3D model vs other 3D ConvNets. Trained and tested on UCF101 split 1. All models were trained from scratch.}
\label{table:TTL}
\end{table}

\subsection{Input Data}

Finding right configuration of input-frames which are fed to the ConvNets for capturing the appearance and temporal information plays a very critical role in temporal ConvNets. For this reason, we start our search by investigating the frame resolutions first, and  we then investigate the frame sampling rate. 

\paragraph{\textbf{Frame Resolution:}}
We use the DenseNet3D-121 for frame resolution study. We evaluate the model by varying the resolution of the input frames in the following set \{($224\times224$), ($112\times112$)\}. Table~\ref{table:frameResolution} presents the accuracy of DenseNet3D-121 trained on inputs with different frame sizes. In the DenseNet3D and later on the T3D setup, the higher frame size of 224px yields better performance. Driven by this observation, we train the final T3D architecture on Kinetics with $224\times224$ to get the best performance.

\begin{table}[hbt] 
\begin{center}
\resizebox{7.0cm}{!} {
\begin{tabular}{ l ||   c c c c c }
\hline
Frame Resolution  & $224\times224$ &$112\times112$ \\
\hline
Accuracy~\%  	& 69.1\% &61.2\% \\
\hline
\end{tabular}}
\end{center}\vspace{-0.25cm}
\caption{Evaluation results of different frame sampling rates for DenseNet3D-121 model. Trained and tested on UCF101 split 1.}
\label{table:frameResolution} 
\end{table}

\paragraph{\textbf{Frame Sampling Rate:}}
The DenseNet3D-121 has been used for evaluations to find the best frame sampling rate in training and testing phase. We evaluate the model by varying the temporal stride of the input frames in the following set \{1, 2 ,4, 16\}. Table~\ref{table:frameSamplingRate} presents the accuracy of DenseNet3D-121 trained on inputs with different sampling rates. The best results is obtained with sampling rate of 2, which we used for other 3D ConvNets also in the rest of experiments: T3D and ResNet3D.

\begin{table}[hbt] 
\begin{center}
\resizebox{7cm}{!} {
\begin{tabular}{ l ||    c c c c }
\hline
Input Stride  & 1  &2 &4 &16\\
\hline
Accuracy~\%  	& 65.2\% &69.1\% &68.3\% &60.5\%\\
\hline
\end{tabular}}
\end{center}\vspace{-0.3cm}
\caption{Evaluation results of different frame sampling rates for DenseNet3D-121 model. Trained and tested on UCF101 split 1.}
\label{table:frameSamplingRate}
\end{table}

\subsection{HMDB51, UCF101, and Kinetics Datasets}

We evaluate our proposed method on three challenging video datasets with human actions, namely HMDB51~\cite{hmdb51}, UCF101~\cite{ucf101}, and Kinetics~\cite{i3d}. Table~\ref{table:databases} shows the details of the datasets. For all of these datasets, we use the standard training/testing splits and protocols provided as the original evaluation scheme. For HMDB51 and UCF101, we report the average accuracy over the three splits and for Kinetics, we report the performance on the validation and test set.

\textbf{\textbf{Kinetics:}}
Kinetics is a new challenging human action recognition dataset introduced by~\cite{i3d}, which contains 400 action classes. There are two versions of this dataset: untrimmed and trimmed. The untrimmed videos contain the whole video in which the activity is included in a short period of it. However, the trimmed videos contain the activity part only. We evaluate our models on the trimmed version. We use the whole training videos for training our models from scratch. Our result for both the DenseNet3D model and the T3D model are reported in the Table~\ref{table:kinetics_results}.

\textbf{\textbf{UCF101:}}
For evaluating our T3D architectures, we first trained them on the Kinetics dataset, and then fine-tuned them on the UCF101. Furthermore, we also evaluate our models by training them from scratch on UCF101 using randomly initialized weights to be able to investigate the effect of pre-training on a huge dataset, such as Kinetics. 

\textbf{\textbf{HMDB51:}}
Same as UCF101 evaluation we fine-tune the models on HMDB51, which were pre-trained from scratch on Kinetics. Similarly, here also we evaluate our models by training them from scratch on HMDB51 using randomly initialized weights.

\begin{table}[hbt] 
\begin{center}
\resizebox{7cm}{!} {
\begin{tabular}{ l |   c     c     c}
\hline
Data-set &  \# Clips &\# Videos &\# Classes\\
\hline
\hline
HMDB51~\cite{hmdb51}  		& 6,766  &	3,312 & 51\\
UCF101~\cite{ucf101}  		& 13,320 &	2,500& 101\\
Kinetics~\cite{i3d}   		& 306,245&	306,245& 400\\
\hline
\end{tabular}}
\end{center}\vspace{-0.3cm}
\caption{Details of the datasets used for evaluation. The `Clips' shows the the total number of short video clips extracted from the `Videos' available in the dataset.}
\label{table:databases}
\end{table}

\subsection{Implementation Details}

We use the PyTorch framework for 3D ConvNets implementation and all the networks are trained on 8 Tesla P100 NVIDIA GPUs. Here, we describe the implementation details of our two schemes, Temporal 3D ConvNets and  supervision transfer from 2D to 3D ConvNets for stable weight initialization.

\noindent
\paragraph{\textbf{Training:}} \emph{}

\noindent
\textbf{$-$ Supervision Transfer: $2D\rightarrow3D$ CNNs.} We employ 2D DenseNet and ResNet architectures, pre-trained on ImageNet~\cite{imagenet},  as used 2D ConvNets. While the 3D CNN is ResNet3D and our T3D network. To the 2D CNN, 32 RGB frames are fed as input. The input RGB images are randomly cropped to a size $224 \times 224$, and then mean-subtracted for the network training. To supervise transfer to the T3D, we replace the previous classification layer of 2D CNN  with a $2$-way softmax layer to distinguish between positive and negative pairs.  We use stochastic gradient descent (SGD) with mini-batch size of 32 with a fixed weight decay of $10^{-4}$ and Nesterov momentum of 0.9. For network training, we start with  learning rate set to 0.1 and manually decrease by a factor of 10 every 30 epochs. The maximum number of epochs is set to 150.

\noindent
\textbf{$-$ Temporal 3D ConvNets.} We train our T3D from scratch on Kinetics. Our T3D operates on a stack of 32 RGB frames. We resize the video to 256px when smaller, and then randomly apply 5 crops (and their horizontal flips) of size $224\times224$.  For network weight initialization, we adopt the same technique proposed in~\cite{densenetweighting}. For the network training, we use SGD, Nesterov momentum of 0.9, weight decay of $10^{-4}$ and batch size of 64.  The initial learning rate is set to 0.1, and reduced by a factor of 10x manually when the validation loss is saturated. The maximum number of epochs for the whole Kinetics dataset is set to 200. Batch normalization also has been applied. We should mention that the proposed DenseNet3D shares the same experimental details as T3D.

\noindent
\paragraph{\textbf{Testing:}}  For video prediction, we decompose each video into non-overlapping clips of 32 frames. The T3D is applied over the video clips by taking a $224\times224$ center-crop, and finally we average the predictions over all the clips to make a video-level prediction.

\subsection{Supervision Transfer} 
To apply our proposed supervision transfer, we have tested 2D ResNet and DenseNet as basic pre-trained on ImageNet, while Res3D and our T3D with randomly initialized using~\cite{densenetweighting}, as target 3D ConvNets.  We show that, a stable weight initialization via transfer learning is possible for 3D ConvNet architecture, which can be used as a good starting model for training on small datasets like UCF101.

Here, we explain the training phase for T3D (see Fig.~\ref{fig:transfer}) case which is similar to the other networks.  To train, we have negative and positive video clip pairs to feed to the networks. Given a pair of 32 frames  and video clip for the same time stamp will go through the 2D DenseNet and T3D. For the 2D network whose model weights are frozen, we do average pooling on the last layer with size of 1024. So, pooled frame features from 2D network are concatenated with clip feature from 3D network ($1024+1024$), and passed to 2 fully connected layers afterward. The fully connected layer sizes are 512, 128. The binary classifier distinguishes between correspondence of negative and positive clip pairs. The T3D network is trained via back-propagation through the network, and the 3D kernels are learned.


Another important aspect of proposing this transfer learning for 3D ConvNets is for finding a cheaper way to train 3D ConvNets when availability of large datasets is at scarce. After pre-training our 3D ConvNets by  described transfer learning, we can use a fraction of a big dataset (e.g. Kinetics) to train the model and still achieve a good performance in fine-tuning on UCF101. In other words, this knowledge transfer reduces the need for more labeled data and very large datasets. As previously mentioned in Sec.~\ref{subsec:tranferlearning}, the transfer learning is done by using approx. 500K unlabeled videos from YouTube8m dataset~\cite{youtube8m}. Since the transfer learning pipeline for 3D ConvNets have been tested with different deep architectures, we clearly show the generalization capacity of our method, and which can be easily adopted for other tasks too. Table~\ref{table:Trasnsfer_result} shows the results, we can observer that via transfer learning we achieve better performance in comparison to training the network from scratch.

\begin{table}[ht] 
\begin{center}
\resizebox{6cm}{!} {
\begin{tabular}{ l | c | c  }
\hline
\textbf{3D ConvNets} &  \textbf{Transfer} & \textbf{FT-Transfer} \\
\hline
ResNet3D 			& 78.2 & 82.1\\ \hline
DenseNet3D			& 79.7 & 82.5\\ \hline
T3D					& 81.3 & 84.7\\ \hline
\end{tabular}}
\end{center} \vspace{-0.3cm}

\caption{Transfer learning results for 3D ConvNets by 2D ConvNets. Both of the results are on UCF101 split1. First column shows the performance of transfered network finetuned directly on UCF101. The second column is finetuned transfered network first on the half of Kinetics dataset and then on UCF101.}
\label{table:Trasnsfer_result}
\end{table}

\begin{table}[hb] 
\begin{center}
\resizebox{7.5cm}{!} {
\begin{tabular}{ l |   c |  c  }
\hline
\textbf{Method} &  \textbf{Top1- Val} & \textbf{Avg-Test}\\
\hline
DenseNet3D 						& 59.5 &  -\\ \hline
\textbf{T3D}					& \textbf{62.2} 	& 71.5\\ \hline
Inception3D 						& 58.9 & 69.7\\ \hline
ResNet3D-38~\cite{3DResHara}		& 58.0 	& 68.9\\ \hline
C3D*~\cite{3DResHara}			& 55.6 	& -\\ \hline
C3D* w/ BN~\cite{i3d}			    & - &  67.8\\ \hline
RGB-I3D w/o ImageNet~\cite{i3d}		& - &  78.2\\ \hline
\end{tabular}}
\end{center} \vspace{-0.3cm}
\caption{Comparison results of our models with other state-of-the-art methods on Kinetics dataset. * denotes the pre-trained version of C3D on the Sports-1M.}
\label{table:kinetics_results}
\end{table}

\subsection{Comparison with the state-of-the-art}

Finally, after exploring and finding an efficient T3D architecture with the best configuration of input-data, we compare our DenseNet3D and T3D with the state-of-the-art methods by pre-training on Kinetics and finetuning on all three splits of UCF101 and HMDB51 datasets. For the UCF101 and HMDB51, we report the average accuracy over all three splits. The results for supervision transfer technique experiments were reported in the previous part of experiments.

Table~\ref{table:kinetics_results} shows the result on Kinetics dataset for T3D compared with state-of-the-art methods. C3D~\cite{c3d} employs batch normalization after each convolutional and fully connected layers (C3D w/ BN), and RGB-I3D which is without pretraining on the ImageNet~(RGB-I3D w/o ImageNet)~\cite{i3d}. The T3D and DenseNet3D achieve higher accuracies than ResNet3D-34, Sports-1M pre-trained C3D and C3D w/ BN which is trained from scratch. However, RGB-I3D achieved better performance which might be the result of usage of longer video clips than ours (64 vs. 32), Although we trained our own version of Inception3D same as I3D~\cite{i3d}, but we could not achieve the same reported performance. As mentioned earlier, due to high memory usage of 3D models we had to limit our model space search and it was not possible to checkout the longer input video clips. Moreover,~\cite{i3d} used larger number of mini-batches by engaging a large number of 64 GPUs that they have used, which plays a vital role in batch normalization and consequently training procedure.

\begin{table}[t] 
\begin{center}
\resizebox{7.5cm}{!} {
\begin{tabular}{ l |   c |  c   }
\hline
\textbf{Method} &  \textbf{UCF101} & \textbf{HMDB51}\\
\hline
DT+MVSM~\cite{dtmvsv}				& 83.5 & 55.9	\\ \hline
iDT+FV~\cite{idt}					& 85.9 & 57.2	\\ \hline
C3D~\cite{c3d}						& 82.3 & 56.8	\\ \hline
Conv Fusion~\cite{pooling}					& 82.6 & 56.8	\\ \hline
Conv Pooling~\cite{n3d} 				 & 82.6 & 47.1	\\ \hline
Spatial Stream-Resnet~\cite{spatialRes} & 82.3 & 43.4	\\ \hline
Two Stream~\cite{twostream}		& 88.6 & $-$	\\ \hline
F$_{ST}$CV (SCI fusion)~\cite{sun3d}	& 88.1 & 59.1	\\ \hline
TDD+FV~\cite{tdd}					& 90.3 &	63.2	\\ \hline
TSN-RGB~\cite{tsn}						& 85.7 & -	\\ \hline
Res3D~\cite{res3d}	& 85.8 & 54.9	\\ \hline
\hline
ResNet3D	& 86.1 & 55.6	\\ \hline
Inception3D	& 87.2 & 56.9	\\ \hline
DenseNet3D	& 88.9 & 57.8	\\ \hline
\hline
\textbf{T3D (ours)}				& \textbf{90.3}		&  \textbf{59.2}	\\ \hline
\textbf{T3D-Transfer (ours)}	& \textbf{91.7}		&  \textbf{61.1}	\\ \hline
\textbf{T3D+TSN (ours)}			& \textbf{93.2}		&  \textbf{63.5}	\\ \hline
\end{tabular}}
\end{center} \vspace{-0.3cm}
\caption{Accuracy (\%) performance comparison of T3D with state-of-the-art methods over all three splits of UCF101 and HMDB51.}
\label{table:state_comparison_UCFHMDB}
\end{table}


Table~\ref{table:state_comparison_UCFHMDB} shows the results on UCF101 and HMDB51 datasets for comparison of T3D with other RGB based action recognition methods. Our T3D and DenseNet3D models outperform the Res3D~\cite{res3d}, Inception3D and C3D~\cite{c3d} on both UCF101 and HMDB51 by 93.2\% and 63.5\% respectively. As mentioned before we trained Inception3D, a similar architecture to the I3D~\cite{i3d} (without using ImageNet) on Kinetics and fine-tuned it on UCF101 and HMDB51 to be able to have a more fair comparison. As shown in the Table~\ref{table:state_comparison_UCFHMDB}, T3D performs better than Inception3D by almost 4\% on UCF101. Furthermore, DenseNet3D and T3D achieve the best performance among the methods using only RGB input on UCF101 and HMDB51. Moreover it should be noted that, the reported result of RGB-I3D~\cite{i3d} pre-trained on ImageNet and Kinetics by Carreira et al.~\cite{i3d} is better than us on both UCF101 and HMDB51, this might be due to difference in usage of longer video clips and larger mini-batch sizes by using 64 GPUs. Furthermore, we note that the state-of-the-art ConvNets~\cite{i3d,tsn} use  expensive optical-flow maps in addition to RGB input-frames, as in I3D which obtains a performance of 98\% on UCF101 and 80\% on HMDB51. However the high cost of computation of such data limits their application at real world large scale applications. As additional experiments, we study the effect of feature fusion methods like TSN~\cite{tsn} on our T3D video features. TSN intends to encode the long term information coming from video clips. We employ the technique of TSN, but here we use our T3D features from 5 non-overlapping clips of each video for encoding via TSN aggregation method. The T3D+TSN results are reported in the same table. This simple feature aggregation method on T3D shows major improvement over using 2D CNN feature extraction from single RGB frames using the same aggregation method. 

Note that, in our work we have not used dense optical-flow maps, and still achieving comparable performance to the state-of-the-art methods~\cite{tsn}. This shows the effectiveness of our T3D to exploit temporal information and capture long-range dynamics in video-clips. This calls for efficient methods like ours instead of computing the expensive optical-flow information (beforehand) which is very computationally demanding, and also difficult to obtain for large scale datasets.

\section{Conclusion} \label{sec:conclusion}

In this work, we introduce a new `Temporal Transition Layer'~(TTL) that models variable temporal convolution kernel depths. We clearly show the benefit of exploiting temporal depths over shorter and longer time ranges over fixed 3D homogeneous kernel depth architectures. In our work, we also extend the DenseNet architecture with 3D convolutions, we name our architecture as `Temporal 3D ConvNets'~(T3D). Our TTL feature-maps are densely propagated throughout and learned in an end-to-end learning. The TTL feature-maps model the feature interaction in a more expressive and efficient way without an undesired loss of information throughout the network. Our T3D is evaluated on three challenging action recognition datasets, namely HMDB51, UCF101, and Kinetics. T3D architecture achieves state-of-the-art performance on HMDB51, UCF101 and comparable results on Kinetics , in comparison to other temporal deep neural network models. Even though, in this paper, we have employed TTL to T3D architecture, our TTL has the potential to generalize to any other 3D architecture too. Further, we show the benefit of transfer learning between cross architectures, specifically supervision transfer from 2D to 3D ConvNets. This provides a valuable and stable weight initialization for 3D ConvNets instead of training it from scratch and this also avoids the computational costs. However, our transfer learning approach is not limited to transfer supervision between RGB models only, as our solution can be easily adopted across modalities too.

\emph{}
\noindent
\textbf{Acknowledgements:} This work was supported by DBOF PhD scholarship, KU Leuven:CAMETRON project, and KIT:DFG-PLUMCOT project. The authors would like to thank Sensifai engineering team.

{\small
\bibliographystyle{ieee}
\bibliography{egbib}
}

\end{document}